\documentclass[10pt,twocolumn,letterpaper]{article}

\usepackage{tikz}
\usepackage{multirow}
\usepackage{caption}
\usepackage{subcaption}
\usepackage{float}
\usepackage{amsmath}
\usepackage{amssymb}
\usetikzlibrary{positioning}
\usepackage{amsmath,amssymb,amsfonts}
\usepackage{algorithm}
\usepackage{algpseudocode}
\usepackage[pagenumbers]{cvpr} 

\definecolor{cvprblue}{rgb}{0.21,0.49,0.74}
\usepackage[pagebackref,breaklinks,colorlinks,allcolors=cvprblue]{hyperref}

\def\paperID{17440}
\def\confName{CVPR}
\def\confYear{2025}

\title{Voxel-Aggregated Feature Synthesis: Efficient Dense Mapping for Simulated 3D Reasoning}

\author{Owen Burns\\
University of Central Florida\\
4000 Central Florida Blvd, Orlando, FL 32816\\
{\tt\small ow446044@ucf.edu}\\
\and
Rizwan Qureshi\\
{\tt\small engr.rizwanqureshi786@gmail.com}\\
}

\begin{document}
\maketitle
\begin{abstract}
We address the issue of the exploding computational requirements of recent State-of-the-art (SOTA) open set multimodel 3D mapping (dense 3D mapping) algorithms and present Voxel-Aggregated Feature Synthesis (VAFS), a novel approach to dense 3D mapping in simulation. Dense 3D mapping involves segmenting and embedding sequential RGBD frames which are then fused into 3D. This leads to redundant computation as the differences between frames are small but all are individually segmented and embedded. This makes dense 3D mapping impractical for research involving embodied agents in which the environment, and thus the mapping, must be modified with regularity. VAFS drastically reduces this computation by using the segmented point cloud computed by a simulator's physics engine and synthesizing views of each region. This reduces the number of features to embed from the number of captured RGBD frames to the number of objects in the scene, effectively allowing a "ground truth" semantic map to be computed an order of magnitude faster than traditional methods. We test the resulting representation by assessing the IoU scores of semantic queries for different objects in the simulated scene, and find that VAFS exceeds the accuracy and speed of prior dense 3D mapping techniques.
\end{abstract}
\section{Introduction}
\label{sec:intro}
Rapid advancements in transformer \cite{vaswani2023attentionneed} based models have brought us closer to realizing the long standing goal of building embodied agents capable of perceiving, planning, and acting autonomously \cite{hu2023generalpurpose}. While limited success has already been achieved in some domain-specific scenarios \cite{hu2023planningoriented}, and Large Language Models have demonstrated surprising zero-shot reasoning capabilities \cite{10.3389/frai.2023.1350306}, open-set 3D perception remains impractical for most such systems.

A truly autonomous system must be able to perceive both semantic and spatial environment information in a joint representation \cite{10.3389/fnhum.2014.00566}, and must be able to do so efficiently \cite{zitkovich2023rt}. Recent research building off of 2D vision language models (VLMs) has taken a sizeable step towards that goal with the creation of open-set 3D mapping methods (dense 3D mapping) which align semantic features with point clouds. The resulting clouds can serve as inputs to 3D VLMs \cite{hong20233dllm} or be used as tools by LLM-based agents \cite{gu2023conceptgraphsopenvocabulary3dscene}.

Approaches to dense 3D mapping typically involve segmenting and embedding a sequence of images which are then fused into a 3D representation. The large number of similar images which must be processed in this way to achieve successful 3D fusion results in a computational load which makes these approaches unsuitable for most real time systems, especially those tasked with exploring unseen environments. Even with modern GPUs, these methods of open-set multimodal 3D mapping (dense 3D mapping) can take upwards of 15 seconds per frame \cite{jatavallabhula2023conceptfusion}.

The computational complexity of dense 3D mapping renders it impractical in many situations it would otherwise prove valuable. In particular, research in agentic cooperation frequently takes place in simulation, where high-fidelity perception and efficient computation are crucial for enabling realistic interaction between agents and their environments. Instead, agents are typically provided with text-based observations that leave them blind to their environment, confounding the results of such studies especially when the agents are in close proximity \cite{mandi2023roco}\cite{yu2023language2r}\cite{ahn2022can}.

To address these shortcomings, we propose Voxel-Aggregated Feature Synthesis (VAFS), a novel approach to efficient dense 3D mapping. Instead of embedding each frame received, we create and embed synthetic views of the different point cloud segments in isolation, and then use voxel aggregation to ensure uniform point distribution. By reducing the computation required to implement dense 3D mapping, VAFS makes this technique viable in a broader set of domains, including research which requires real time updates. In particular, we focus on simulator-based applications as such environments readily provide segmented point clouds, further increasing the efficiency gain over fusion-based methods.  Our main contributions are as follows:
\begin{itemize}
\item A novel method to generate synthetic views of regions of interest in a point cloud.
\item A computationally efficient approach to dense 3D mapping capable of uniquely leveraging information available in simulated environments to create ground truth semantic observations for agentic research.
\end{itemize}
\section{Related Works}
\label{sec:related}
\subsection{Multi-Modality and Perception}

Multi-modal perception has seen rapid advancement in recent years. In 2D, models such as CLIP \cite{radford2021clip} and ALIGN \cite{jia2021scalingalign} emerged to encode text and images into a shared latent space, while segmentation models such as SAM \cite{kirillov2023segmentsam} and MASKFORMER \cite{cheng2021perpixelmaskformer} emerged to perform class-agnostic image segmentation.  Semantic segmentation models such as OpenSeg~\cite{ghiasi2022scalingopenseg} and LSeg \cite{li2022languagedrivenlseg} merge those tasks, creating embeddings for objects within images, while captioning models like Blip \cite{li2023blip2} go from image to text. Another trend has been integrating image understanding into pre-existing models, with nearly all major LLM providers such as OpenAI \cite{openai2024gpt4} and Google \cite{team2023gemini} offering native image support and open source methods coming along not long after \cite{awadalla2023openflamingo}\cite{li2023blip2}. More complex tasks like motion and object tracking \cite{li2022panoptic}\cite{carion2020endtoend}\cite{jia2021multiagent}\cite{li2022maskdino} and 3D scene representations \cite{peng2023openscene}\cite{hong20233dllm} have begun to be considered as well as the frontier expands. However, the considerable cost associated with these techniques, attention's $N^2$ runtime in particular, has motivated a number of papers to explore alternatives to representing these modalities in full  or ways to isolate and ignore regions of input lacking in information \cite{zhu2021deformable}\cite{gu2023conceptgraphs}.

\subsection{Dense 3D Mapping}
Dense 3D mapping algorithms\footnote{Rigidly defined in \ref{sec: problemsspec}} generally follow these steps:
\begin{enumerate}
    \item Capture multi-view depth images covering the whole scene
    \item Compute pixel-wise features for each depth image
    \item Fuse the depth images into a 3D environment representation, using some method to combine the features of two points of the fusion operation decides to merge their associated points.
\end{enumerate}
In~\cite{ha2022semantic}, 2D semantic relevancy maps are computed and projected to 3D using the RGB-D depth data, and subsequently used to compute the volume of the relevant region. Other approaches \cite{mazur2022featurerealistic}\cite{Kerr2023LERFLE}\cite{Tschernezki2022NeuralFF}\cite{Liao2024OVNeRFON} extract 2D pixel aligned features with off-the-shelf models and perform 3D fusion with a neural field (NeRF) \cite{sucar2021imap}. NeRF-based methods typically do not create a standalone 3D representation, resulting in slower semantic queries, though Gaussian splatting has been demonstrated to be a viable means of overcoming this limitation \cite{qin2024langsplat3dlanguagegaussian}. The most common setup, detailed in ConceptFusion \cite{jatavallabhula2023conceptfusion}, uses simultaneous localization and mapping (SLAM) to fuse the 2D embeddings \cite{Chen2022OpenvocabularyQS}\cite{Shafiullah2022CLIPFieldsWS}\cite{Huang2022VisualLM}. OV3D \cite{Liu20233DOS} takes this a step further, using text descriptions of image segments created by vision-language models instead of image embeddings to incorporate additional context. In all such setups, large overlap is needed between input images to ensure the fusion step (whether that be direct reconstruction, SLAM~\cite{taketomi2017visual}, or a NeRF~\cite{mildenhall2020nerf}) can achieve a well-aligned representation. This overlap makes pre-processing the image set prohibitively expensive, and limits the usefulness of these methods for systems requiring frequently updated mappings. 
\section{Voxel Aggregated Feature Synthesis}
\label{sec:method}
\subsection{Problem Specification}
\label{sec: problemsspec}
We define dense 3D mapping algorithms to be an algorithms which take some sequence of observations $T\in\mathcal{T}$ (typically depth images) from the environment $\mathcal{S}$ and constructs a 3D representation of the environment $M\in\mathcal{M}$ where each point is associated with a semantic feature relevant to that point. Concretely,
\begin{equation}
\begin{aligned}
\mathcal{T}\in\{(v_1, v_2,...,v_t)|\forall v_t \in T \exists v_i \in S(t) s.t. o_t\equiv v_i\} \\
\mathcal{M}=\{\{\bar{p_j},\mathbf{F_j}\}|\bar{p_j}\in\mathbb{R}^3,\bar{F_j}\in\mathbb{R}^N,1j\in\mathcal{S}_p\} \\
f: T\rightarrow M
\end{aligned}
\end{equation}
We take inspiration from ConceptFusion's definition of dense 3D mapping problems, but relax the requirement that $v$ must be a depth image and 
that $\mathcal{M}$ must include normals and confidence counts to arrive at the above. These constraints are unnecessary in this context because the simulator is capable of giving us a ground truth point cloud, removing the need for us to calculate camera poses or keep track of an estimate of position error. Unfortunately, this does mean adding new constraints on the simulation environment, which are detailed in \ref{sec: Assumptions}.

\subsection{Assumptions}
\label{sec: Assumptions}
We assume that for simulation environment $\mathcal{S}$ with points $\mathcal{S}_p$ and objects $\mathcal{S}_o$ at time step $t$, $\mathcal{S}$ is capable of providing a coordinate $p$, color $r$, and object reference $i$ for each point $k$ in the ground truth point cloud $\mathcal{P}$ as follows.
\begin{equation}
\begin{aligned}
\mathcal{P}=\{\{\bar{p}_k, \bar{r}_k, o_k\}|\bar{p}, \bar{r}\in\mathbb{R}^3, k\in\mathcal{S}_p, o\in\mathcal{S}_o\} \\
f_p: (\mathcal{S}, t, k)\rightarrow\{\bar{p_k}, \bar{r_k}, i_k\}
\end{aligned}
\end{equation}
We also assume that:
\begin{itemize}
    \item  All points in a given object have the same semantic meaning in the context of what information is being encoded in the pixels.
    \item  The information encoded in a pixel is dependent only upon the object of which that pixel is a part.
    \item We have a model that can transform a view into its corresponding 2D feature defined as $f_e:v^m\rightarrow e^m, m\in\{G,o\}$.
\end{itemize}

In our specific implementation, we use the Mujoco simulator following after RoCo \cite{mandi2023roco}, and define cameras in the simulator's XML file to follow each object in the environment at a fixed distance and angle.

\begin{figure*}[h]
    \centering
    \input{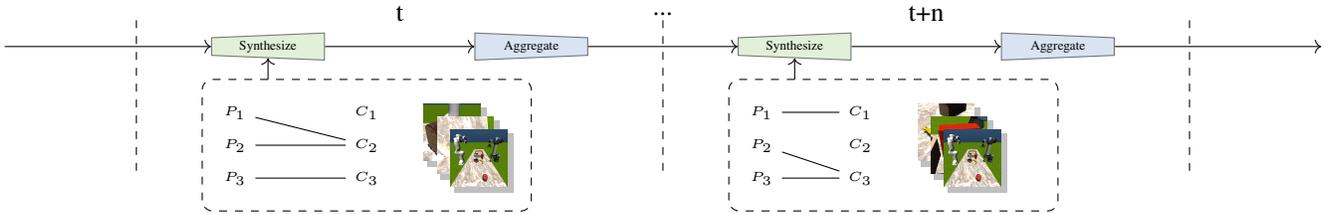}
    \caption{The high-level workflow of VAFS. At each time step, we associate points $P$ with segments $C$ and render views of the regions of interest. We then align embeddings of those views with the point cloud and run voxel aggregation to ensure the distribution of points remains uniform. Subsequent time steps represent updates to the point cloud, and the process runs again with new views generated for segments of the point cloud that have changed.}
    \label{fig:your-label-here}
\end{figure*}

\subsection{Algorithm}

\textbf{Simulation query}: At each time step $t$, we query the environment for current point cloud $\mathcal{P}_t$.

\begin{equation}
    \begin{aligned}
\mathcal{P}_t=\{f_p(\mathcal{S},t,k)|k\in\mathcal{S}_p\} \\
    \end{aligned}
\end{equation}

\textbf{View Synthesis}: We then group the points by their object reference, and render a synthetic view $v^o$ of the object corresponding to those points by aligning a camera with the average estimated normal of the points. We repeat the process for the entire set of points to get the global view $v^G$. This process is detailed in Algorithm~\ref{alg:viewsynth}.

\begin{algorithm}
\caption{View Synthesis}
\label{alg:viewsynth}
\begin{algorithmic}
\State $\mathcal{O} \gets \{\bar{p_k}, \bar{r_k}|f_p(\mathcal{S},t,k)[2] = o, \bar{p_k}\ is\ not\ an\ outlier\}$
\State $\mathcal{N} \gets GET\_NORMALS(O)$
\State $\mathcal{N} \gets \frac{1}{|\mathcal{N}|} \sum_{n \in \mathcal{N}} n$
\State $\mathcal{N} \gets \frac{\mathcal{N}}{\|\mathcal{N}\|}$
\State $elevation \gets \arcsin\left(\mathcal{N}_z\right)$
\State $azimuth \gets \arctan\left(\frac{\mathcal{N}_y}{\mathcal{N}_x}\right)$
\State $\mathcal{V} \gets RENDER(O, elevation, azimuth)$
\State \Return $\mathcal{V}$
\end{algorithmic}
\end{algorithm}

\textbf{Local feature computation}: We then compute the set of 2D object features $\mathcal{E}_t$ and the the 2D global feature ${e^G}_t$. 
\begin{equation}
    \begin{aligned}
\mathcal{E}_t=\{f_e(v^o)|v^o\in\mathcal{T}_t\} \\
{e^G}_t=f_e(v^G)
    \end{aligned}
\end{equation}

In order to contextualize the object features, we follow the method of ConceptFusion~\cite{jatavallabhula2023conceptfusion} to compute the importance of an object to the scene by assessing its difference from the global feature. We then use this to compute a weighted sum aggregating information from that object feature and the global feature and normalize it to get the point cloud feature ${c^o}_t$. Then, we assign these features to their corresponding points in $\mathcal{P}_t$ to get the update concept cloud as follows:
\begin{equation}
{\mathcal{M}^u}_t=\{\{\bar{p},{c^o}_t\}|\bar{p}\in\mathcal{P}_t,o\in\mathcal{P}_t\Rightarrow {c^o}_t\triangleq o\}
\end{equation}

\textbf{Voxel aggregation}: We use voxel pooling to both maintain a consistent density of points in our concept cloud and ensure that relative positioning relationships at the borders between objects are represented explicitly. Using $\nu$ as the voxel size and $\epsilon$ as the increment (all outcomes in \ref{sec: Results} used 0.1 for the starting size and increment), we create a voxel grid $\mathcal{V}$ starting from the origin and for each voxel $\mathcal{V}_{ijk}$ add and normalize the features of all the points within to create the combined feature and take their centroid to be the new position.

\begin{equation}
    \begin{aligned}
\mathcal{V}_{ijk}=\{{\mathcal{M}^u}_{ti}|\lfloor\frac{{{\mathcal{M}^u}_{ti}}_{\bar{p}}}{\nu}\rfloor=(i,j,k)\} \\
\rho \leftarrow \frac{1}{|\mathcal{V}_{ijk}|}\sum_{x\in\mathcal{V}_{ijk}}x_p \\
\phi \leftarrow ||\sum_{x\in\mathcal{V}_{ijk}}x_c||_2 \\
{\mathcal{M}_t}_{ijk}\leftarrow\{\rho,\phi\}
    \end{aligned}
\end{equation}
\section{Experimental Setup and Results}
\label{sec: Results}

\begin{figure*}[h]
    \includegraphics[width=\textwidth]{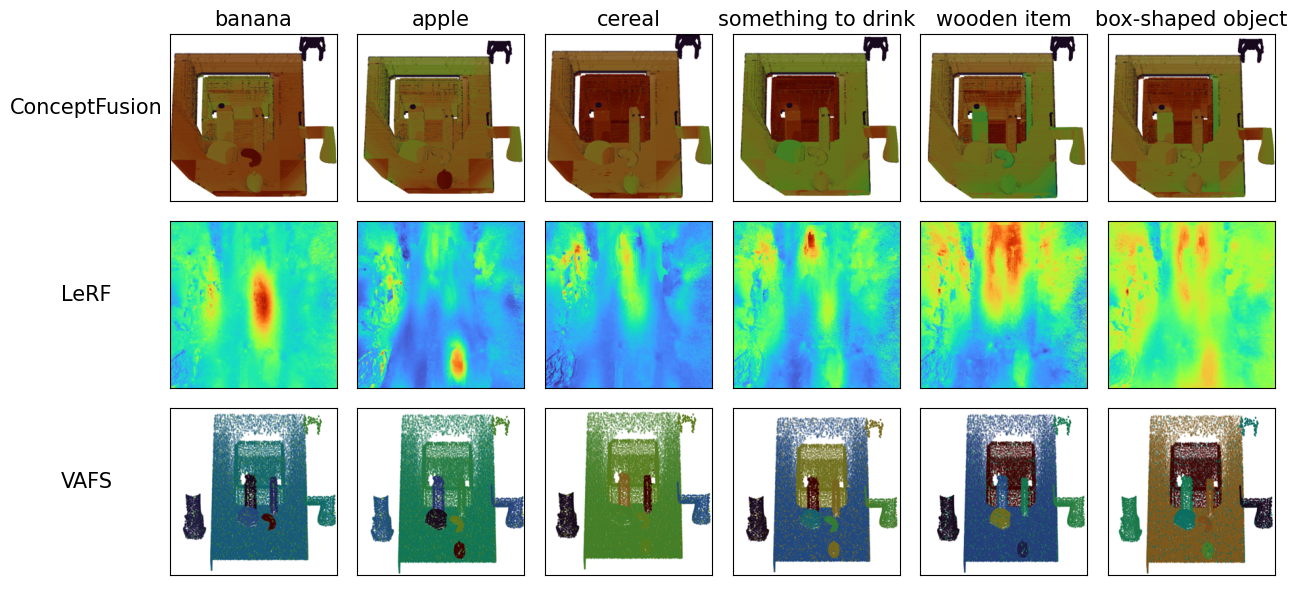}
    \caption{Relevancy maps for semantic queries.}
    \label{fig:semquery}
\end{figure*}

One of the most common use cases for 3D dense mapping models is creating 3D semantic representations of an environment which can be queried in various modalities. To validate the performance of VAFS, We compared it to ConceptFusion~\cite{jatavallabhula2023conceptfusion} and LeRF~\cite{Kerr2023LERFLE} on the task of identifying regions in a simulated scene (we used a scene from the RoCoBench dataset \cite{mandi2023roco}) that were most strongly related to the query. We found that VAFS computed the map far faster than either baseline (Table~\ref{fig:times}), while maintaining a higher IoU score across the board even on indirect semantic queries (Table~\ref{fig:scores}). We also found that our synthetic view generation improved the localization abilities of VAFS compared to the fusion-based baselines. Both baselines gave additional relevance to the regions around the target region, while VAFS prevented this blurring of features (Fig.~\ref{fig:semquery}).



\begin{table*}[h]
    \centering
    \begin{tabular}{ |p{3cm}||p{3cm}|p{3cm}|p{3cm}|  }
     \hline
     \multicolumn{4}{|c|}{Runtime (s)} \\
     \hline
     Method       &2D feature computation&3D Fusion&Total \\
     \hline
     ConceptFusion&1279                    &257     & 1536      \\
     \hline
     LERF  &791&5040&5831 \\
     \hline
     VACC (Ours)  &\textbf{175}&\textbf{145}&\textbf{189} \\
     
     \hline
    \end{tabular}
    \caption{Runtime on a single Nvidia L4, processing 236 images from our simulation environment.}
    \label{fig:times}
\end{table*}

\begin{table*}[h]
    \centering
    \begin{tabular}{ |p{2.5cm}||p{1cm}|p{1cm}|p{1cm}|p{1.5cm}|p{1cm}|p{1cm}|  }
     \hline
     \multicolumn{7}{|c|}{Text query IoU on Simulator Environment} \\
     \hline
     Method       &banana&apple&cereal&something to drink&wooden item&box-shaped object\\
     \hline
     ConceptFusion  &0.132&0.496&0.016&0.030&0.649&\textbf{0.662} \\
     \hline
     LeRF  &0.444&0.273&0.0&0.182&0.273&0.284 \\
     \hline
     VACC (Ours)  &\textbf{0.790}&\textbf{0.656}&\textbf{0.544}&\textbf{0.733}&\textbf{0.713}&0.523 \\
     \hline
    \end{tabular}
    \caption{The IoU scores of VACC versus ConceptFusion and LeRF.}
    \label{fig:scores}
\end{table*}
\section{Conclusion}
We present VACC as a computationally efficient way to implement dense 3D mapping algorithms in simulation, increasing the accessibility of these algorithms to simulation-based agentic research. We demonstrate an order of magnitude decrease in runtime while achieving better performance than fusion-based approaches. Future lines of work will include extending VAFS to include point cloud segmentation and testing the method on real-world footage.
{
    \small
    \bibliographystyle{ieeenat_fullname}
    \bibliography{main}
}


\end{document}